\documentclass[cameraready]{Interspeech}
\usepackage{multirow}
\usepackage[normalem]{ulem}
\useunder{\uline}{\ul}{}
\usepackage{hyperref}
\usepackage{xurl}

\title{Multimodal LLMs are not all you need for Pediatric Speech Language Pathology}

\author[affiliation={1}, orcid=0009-0006-3607-349X, equalcontribution,correspondingauthor]{Darren}{Fürst}
\author[affiliation={1}, orcid=0000-0001-8141-5264, equalcontribution]{Sebastian}{Steindl}
\author[affiliation={1}, orcid=0000-0003-2649-3731]{Ulrich}{Schäfer}

\address{
    $^1$ Ostbayerische Technische Hochschule Amberg-Weiden
}

\email{d.fuerst@oth-aw.de, s.steindl@oth-aw.de, u.schaefer@oth-aw.de}


\usepackage{xcolor}
\usepackage{xurl}

\usepackage{comment}

\usepackage[nolist]{acronym}
\begin{document}
\begin{acronym}
    \acro{LLM}{Large Language Model}
    \acro{SOTA}{State-of-the-Art}
    \acro{SRM}{Speech Representation Model}
    \acro{ASR}{Automatic Speech Recognition}
    \acro{SLP}{Speech Language Pathology}
    \acro{SSD}{Speech Sound Disorders}
\end{acronym}

\maketitle

\begin{abstract} 
\ac{SSD} affect roughly five percent of children, yet speech-language pathologists face severe staffing shortages and unmanageable caseloads. We test a hierarchical approach to SSD classification on the granular multi-task SLPHelmUltraSuitePlus benchmark. We propose a cascading approach from binary classification to type, and symptom classification. By fine-tuning \acp{SRM}, and using targeted data augmentation we mitigate biases found by previous works, and improve upon all clinical tasks in the benchmark. We also treat \ac{ASR} with our data augmentation approach. Our results demonstrate that \acp{SRM} consistently outperform the LLM-based state-of-the-art across all evaluated tasks by a large margin. 
We publish our models and code to foster future research\footnote{Code available on \url{https://github.com/N0tAScooby/Multimodal-LLMs-are-not-all-you-need-for-Pediatric-Speech-Language-Pathology}.}.

\end{abstract}

\section{Introduction}
Studies suggest that roughly five percent of children are affected by \ac{SSD} \cite{shriberg1999prevalence,black2015communication}. 
\ac{SSD} have been shown to increase the risk of social, academic, and emotional challenges for affected children, also during interaction with peers \cite{hitchcock2015social,daniel2017children,foster2023speech}.
\ac{SSD} during childhood can have long-lasting negative effects even during adulthood \cite{mccormack2009systematic}.
Research supports the benefit of early therapeutic intervention \cite{mcis1998effectiveness,jones2005randomised}.

However, speech-language pathologists face increasingly unmanageable caseloads and severe staffing shortages~\cite{asha2024, ashaSchool2024}.

Therefore, an urgent gap exists and advances in Machine Learning might provide a technical solution \cite{patel2025sound}. Research into Machine Learning for disordered speech started with a focus on ASR, where systems achieved much lower performance compared to typical speech, with efforts like Project Euphoria aiming to provide the required training data \cite{macdonald2021disordered}. 
The collection of child speech data is even more challenging due to privacy constraints and behavioral challenges \cite{hansenBestPracticesConsiderations2025}.
The UltraSuite \cite{eshkyUltraSuiteRepositoryUltrasound2018} dataset is a notable resource for child speech data.
Recently, Patel et al. \cite{patel2025sound} published a dataset that combines audio samples from multiple previous publications, including Ultrasuite, providing expert transcription and fine-grained annotations for disorder type and symptom, resulting in a new benchmark for \ac{SSD} classification and \ac{ASR} of disordered speech.

\begin{figure}[!t]
    \centering
    \includegraphics[width=\linewidth]{imgs/Hierarchical_Task.png}
    \caption{Hierarchical approach proposed in this work. Only samples that are classified as pathological speech are routed to the T2 and T3 classifier, where the annotation becomes more detailed.}
    \label{fig:hierarchical_approach}
\end{figure}

Previous works have addressed these tasks for speech pathologies in children.
For example, Ng et al. \cite{ngAutomaticDetectionSpeech2022} propose a subject-level \ac{SSD} detection system based on speaker representations in Cantonese child speech. 
Later, Kim et al.~\cite{kim2024automatic} build an \ac{SSD} detection for Korean child speech, and Marx et al.~\cite{marx2025automatic} for German child speech.
Marx et al.~\cite{marx2025automatic} additionally treat \ac{ASR} of disordered speech, as do Rosero et al.~\cite{roseroAdvancingPediatricASR2025} and earlier Getman et al.~\cite{getman2022wav2vec2} and Smith et al.~\cite{smithImprovingChildSpeech2017}. To mitigate data scarcity, literature has proposed to, e.g., use transfer learning from adult speech \cite{smithImprovingChildSpeech2017,getman2022wav2vec2}, and synthetically generated audio \cite{roseroAdvancingPediatricASR2025}.

\begin{table*}[t]
\centering
\caption{Main results of the hierarchical classification pipeline. $\Delta$ F1 denotes the absolute mean Macro F1 improvement compared to the ablation study. $\ast$: As reported by \cite{patel2025sound}.}
\resizebox{\textwidth}{!}{%
\begin{tabular}{lcccccccc}
\toprule
Model & \multicolumn{2}{c}{T1} & \multicolumn{3}{c}{T2} & \multicolumn{3}{c}{T3} \\
 & F1 ($\uparrow$) & Recall ($\uparrow$) & F1 ($\uparrow$) & $\Delta$ F1 & Recall ($\uparrow$) & F1 ($\uparrow$) & $\Delta$ F1 & Recall ($\uparrow$) \\ \midrule
WavLM-large & \textbf{0.956 $\pm$ 0.019} & \textbf{0.956 $\pm$ 0.020} & \textbf{0.697 $\pm$ 0.021} & +0.140 & \textbf{0.702 $\pm$ 0.029} & \textbf{0.354 $\pm$ 0.027} & +0.034 & \textbf{0.391 $\pm$ 0.040} \\
wav2vec2-base & 0.797 $\pm$ 0.078 & 0.804 $\pm$ 0.075 & 0.590 $\pm$ 0.000 & +0.107 & 0.618 $\pm$ 0.000 & 0.336 $\pm$ 0.039 & +0.043 & 0.361 $\pm$ 0.034 \\
wav2vec2-large & 0.889 $\pm$ 0.033 & 0.890 $\pm$ 0.032 & {\ul 0.646 $\pm$ 0.051} & +0.297 & 0.655 $\pm$ 0.047 & {\ul 0.351 $\pm$ 0.047} & +0.096 & {\ul 0.387 $\pm$ 0.032} \\
Hubert large & {\ul 0.936 $\pm$ 0.025} & {\ul 0.935 $\pm$ 0.026} & 0.623 $\pm$ 0.029 & +0.024 & {\ul 0.658 $\pm$ 0.021} & 0.333 $\pm$ 0.023 & +0.047 & 0.377 $\pm$ 0.020 \\ \midrule
Phi-4-multimodal-instruct$\ast$ & 0.535 $\pm$ 0.016 & - & 0.163 $\pm$ 0.006 & - & - & 0.118 $\pm$ 0.010 & - & - \\
gpt-4o-transcribe$\ast$ & 0.373 $\pm$ 0.014 & - & 0.318 $\pm$ 0.017 & - & - & 0.201 $\pm$ 0.019 & - & - \\
whisper-gpt4o$\ast$ & 0.391 $\pm$ 0.013 & - & 0.245 $\pm$ 0.013 & - & - & 0.222 $\pm$ 0.015 & - & - \\ \bottomrule
\end{tabular}%
}
\label{tab:main}
\end{table*}

In this paper, we study whether the transition to (multimodal) \ac{LLM} for speech pathology tasks \cite{patel2025sound} is justified by performance gains, or if traditional audio-specific \ac{SRM} still provide value. We propose a hierarchical classification framework that leverages the hierarchical structure within different \ac{SSD} classification tasks.  
Additionally, we use data augmentation to mitigate gender bias and class imbalance.
Our results show that fine-tuned \acp{SRM} outperform the LLM-based SOTA \cite{patel2025sound} in all four evaluated tasks by a large margin. This is true not only for the hierarchical classification, but also for our ablation study.

Our work treats three main research questions (RQs).

\textbf{RQ1:} Can \acp{SRM} outperform (multimodal) \ac{LLM} on speech pathology classification and \ac{ASR}?

\textbf{RQ2:} Can the hierarchical annotation structure be leveraged to improve the classification?

\textbf{RQ3:} Can data augmentation mitigate gender bias?

Our main contributions are that we (i) propose a hierarchical classification architecture, leveraging the natural annotation structure, (ii), show that \ac{SRM} consistently outperform LLM-based approaches on all evaluated tasks, improving the \ac{SOTA}, and (iii) show that pitch-based data augmentation can mitigate gender bias.
We publish our code and trained models to foster future research\footnote{ Trained models available on \url{https://huggingface.co/N0tAScooby/Multimodal-LLMs-are-not-all-you-need-for-Pediatric-Speech-Language-Pathology}}.

% ++++++++++++++++++++++++++++++++++++++++++++++++++++++++++++++
\section{Method and Materials}
% ++++++++++++++++++++++++++++++++++++++++++++++++++++++++++++++

Our work evaluates the performance of three widely used architectures for speech representation to modern (multimodal) \ac{LLM} for speech pathology classification.
As \acp{SRM} we utilize the architectures Hubert \cite{hubert}, wav2vec2 \cite{wav2vec}, and WavLM \cite{wavlm}. 
We compare these to the best performing models from \cite{patel2025sound}, which are based on the architectures of Phi-4 \cite{phi4}, GPT-4o \cite{gpt4o}, and Whisper \cite{whisper}.

We perform our comparison on the SLPHelmUltraSuitePlus dataset~\cite{patel2025sound}, which was proposed as a benchmark across clinical, pediatric \ac{SLP}.
It contains $926$ audio child speech samples from $262$ unique speakers and has fine-grained, manual annotations validated by trained speech-language pathologists.
This benchmark includes three classification tasks (T1, T2, T3).
The first task is \textbf{Disorder Diagnosis} (T1). T1 is the binary classification between typical and disordered speech.
Next, the \textbf{Disorder Type Classification} (T2), differentiates between \textit{articulation disorders}, such as lisps, and \textit{phonological disorders}, such as partial omissions. 
Finally, the \textbf{Disorder Symptom Classification} (T3) is the most granular task, where the model should classify which symptoms are present. The annotation for T3 includes the symptoms \textit{Addition, Substitution, Omission} and \textit{Stuttering}. 

Additionally to these three classification tasks, we also treat \ac{ASR}, using an array of Whisper models~\cite{whisper}. The dataset includes ground truth transcriptions from a trained speech pathologist. While automated transcription is highly valuable in clinical practice, \ac{LLM} tend to automatically over-correct non-normative speech \cite{ahn2025automatic}. We demonstrate that fine-tuning \ac{SRM} on the dataset's pathologist-provided transcriptions \cite{patel2025sound} yields vastly better \ac{ASR} results than fine-tuned \ac{LLM}.
For a more detailed description of the dataset and task, we refer the interested reader to the original publication of the dataset \cite{patel2025sound}. 

\subsection{Hierarchical Classification}
Within the three classification tasks exists a natural hierarchy in the annotations. The first task is a binary classification between typical and pathological speech. T2 and T3 concern classifying pathological speech in more detail, regarding its type and determining symptoms. 
As described in \autoref{sec-augmentation}, while the dataset is balanced for the binary task T1, the labels for the consecutive tasks are highly imbalanced. Given this and the strong performance of classifiers on T1 in preliminary experiments (Macro F1 of more than 0.9 on the validation data), we propose to leverage the hierarchical structure. We implement a cascading pipeline where a model trained on T1 first decides whether a speech sample is pathological. Subsequently, a downstream classifier for T2 or T3 classifies the type or symptom for these pathological samples only, allowing the downstream classifiers to learn to decide between pathologies only. This greatly reduces the data imbalance, aiding in training stability and model performance.
A visualization of this pipeline is shown in \autoref{fig:hierarchical_approach}.

\subsection{Data Augmentation}\label{sec-augmentation}

We employ two types of augmentation, aimed at improving \ac{ASR} and the three classification tasks, as well as mitigating the gender bias found by Patel et al.~\cite{patel2025sound}:

\textit{Gaussian Noise}: Gaussian Noise is added with a parameterizable maxmimum amplitude to simulate noisier environments.  This also decrease potential biases in the recording environment or microphone, necessitating the model to learn more robust features. Gaussian noise is applied with probability $p_N$.

\textit{Pitch Shifting}: Pitch Shifting is employed to decrease gender bias in models identified by Patel et al. \cite{patel2025sound}  with lower performance for female speakers. Further, this introduces variability in pitch ranges for the different pathologies, decreasing bias to speakers with specific pathologies. The maximum and minimum pitch shifts are parameterized for the trainings.  Pitch-shifting augmentation is applied with probability $p_G$ and gender-stratified.
Furthermore, we employ oversampling, with the oversampling factor being a tunable hyperparameter, and apply weighting to the cross-entropy loss to mitigate the high data imbalance. While the samples for T1 are balanced, for T2 the two majority classes make up for $91\,\%$ of data, whereas for T3, the two majority classes, form $72\,\%$. 

% ++++++++++++++++++++++++++++++++++++++++++++++++++++++++++++++
\section{Experimental Setup}
% ++++++++++++++++++++++++++++++++++++++++++++++++++++++++++++++
\begin{table*}[!ht]
\caption{Fine-tuned \ac{ASR} models. Best in \textbf{bold}, second best {\ul underlined}. $\ast$: As reported in \cite{patel2025sound}.}
\centering
\resizebox{0.9\textwidth}{!}{%
\begin{tabular}{llcccccc}
\toprule
 & \textbf{Model} & \textbf{EM ($\uparrow$)} & \textbf{F1 ($\uparrow$)} & \textbf{WER ($\downarrow$)} & \textbf{WIP ($\uparrow$)} & \textbf{MER ($\downarrow$)} & \textbf{CER ($\downarrow$)} \\ \hline
\multirow{3}{*}{\rotatebox[origin=c]{90}{\shortstack{Pure\\Transcr.}}} 
& Whisper-large-v2 & {\ul $0.607 \pm 0.052$} & {\ul $0.793 \pm 0.032$} & {\ul $0.260 \pm 0.155$} & {\ul $0.666 \pm 0.073$} & {\ul $0.228 \pm 0.082$} & \textbf{0.210 $\pm$ 0.140} \\
& Whisper-large-v3 & $0.520 \pm 0.019$ & $0.748 \pm 0.011$ & $0.369 \pm 0.118$ & $0.563 \pm 0.065$ & $0.314 \pm 0.071$ & $0.330 \pm 0.149$ \\
& Whisper-large-v3-turbo & \textbf{0.640 $\pm$ 0.051} & \textbf{0.814 $\pm$ 0.033} & \textbf{0.194 $\pm$ 0.031} & \textbf{0.708 $\pm$ 0.039} & \textbf{0.187 $\pm$ 0.028} & $0.365 \pm 0.651$ \\ \midrule
\multirow{3}{*}{\rotatebox[origin=c]{90}{\shortstack{Incl. \\Vowel\\Descr.}}} & Whisper-large-v2 & 0.555 $\pm$ 0.037 & 0.703 $\pm$ 0.019 & 0.293 $\pm$ 0.016 & 0.588 $\pm$ 0.020 & 0.273 $\pm$ 0.014 & $0.295 \pm 0.054$ \\
 & Whisper-large-v3 & 0.414 $\pm$ 0.067 & 0.596 $\pm$ 0.065 & 0.495 $\pm$ 0.290 & 0.456 $\pm$ 0.110 & 0.400 $\pm$ 0.118 & $0.392 \pm 0.162$ \\
 & Whisper-large-v3-turbo & 0.553 $\pm$ 0.033 & 0.710 $\pm$ 0.024 & 0.303 $\pm$ 0.018 &  0.584 $\pm$ 0.025 & 0.278 $\pm$ 0.016 & {\ul $0.252 \pm 0.022$} \\ \midrule
\multirow{3}{*}{\rotatebox[origin=c]{90}{\shortstack{LLM\\SOTA}}} & Qwen2-Audio-7B-Instruct* & - & - & 0.572 $\pm$ 0.030 & 0.547 $\pm$ 0.014 & 0.385 $\pm$ 0.013 & - \\
 & Qwen2.5-Omni-3B* & - & - & 0.996 $\pm$ 0.048 & 0.360 $\pm$ 0.013 & 0.569 $\pm$ 0.013 & - \\
 & Qwen2.5-Omni-7B* & - & - & 1.762 $\pm$ 0.215 & 0.440 $\pm$ 0.013 & 0.489 $\pm$ 0.013 & - \\ \bottomrule
\end{tabular}
}
\label{tab:ASR}
\end{table*}

We use the SLPHelmUltraSuitePlus~\cite{patel2025sound} benchmark to evaluate the model performance. 

For the \ac{ASR} task, we fine-tune models using Low-Rank-Adaption (LoRA)~\cite{hu2021loralowrankadaptationlarge}. The models used are: whisper-large-v2, whisper-large-v3, whisper-large-v3-turbo. We fine-tune whisper models, as we hypothesize these to fare better on the specialized \ac{ASR} for this clinical setting and compare their performance to the fine-tuned \acp{LLM} from Patel et al.~\cite{patel2025sound}.
For the classification tasks T1, T2, and T3, we employ a direct audio-classification pipeline by fine-tuning \acp{SRM}.
As \acp{SRM}, we use the pre-trained checkpoints WavLM-large, wav2vec2-base, wav2vec2-large, and Hubert-large-ll60k, motivated by architectures used in prior work.
We freeze the convolutional encoder of the models during fine-tuning to leverage the learnt feature extractions, only tuning the encoder blocks and the classification head.
For training, a weighted cross-entropy loss is used as explained in \autoref{sec-augmentation}, we also experimented with using focal loss, it was however not proven beneficial over a weighted cross-entropy loss.
We perform a stratified train-val-test-split (with $64,\:16,\:20$ percent, respectively), avoiding speaker-leakage across sets. This results in a test-set consisting of $53$ unique speakers. We also make sure that all classes and genders for each task are represented adequately in each of the sets.
For computational efficiency, we trim the maximum audio duration to twelve seconds, which is twice the median duration across the dataset. This is only done for T1-3, not for \ac{ASR}.

Aiming to improve the performance, we conduct a bayesian hyperparameter tuning before we perform the final fine-tuning. The hyperparameters we optimize, and their ranges, are described in \autoref{sec-hyperparams}. We select the best hyperparameters based on the highest Macro F1 score on the validation split\footnote{Best hyperparameters published with code.}. We conduct the hyperparameter search for every architecture independently. 
Using the best hyperparameters, we run the fine-tuning and evaluation on the test data split ten times with different random seeds. We report our results as the $\mu \pm \sigma$ across these ten runs. We train for a maximum of 20 epochs and use the best checkpoint based on the validation dataset.

For the hierarchical classification, we choose the T1-Classifier that achieved the highest validation-set F1-Macro score, which is the WavLM model. We split the dataset as previously described, before filtering the train-validation-test-sets each separately to contain pathological speech only. Thus, we ensure no data leakage occurs and speakers remain the same as for regular T1-3 classification, providing comparability between approaches. The models for T2 and T3 are then trained on this subset that contains only true instances of pathologies.

We compare our results to the current \ac{SOTA} on the benchmark we use \cite{patel2025sound}. Patel et al. \cite{patel2025sound} evaluate numerous (multimodal) \ac{LLM} in two inference pipelines. First, they input the audios combined with the prompt to describing the classification task into a multimodal \ac{LLM}. In the second method, they first transcribe the audio with an \ac{ASR} model, and then embed the transcribed text into the classification prompt for a text-only \ac{LLM}. 
For each task, we compare to their best performing model according to the Macro F1 score.

\begin{table}[b]
\caption{Gender-stratified Macro F1. $\ast$: As reported by \cite{patel2025sound}.}
\centering
\resizebox{\columnwidth}{!}{
\begin{tabular}{lcccc}
\toprule
Model & \multicolumn{1}{l}{Gender} & T1 ($\uparrow$) & \multicolumn{1}{c}{T2 ($\uparrow$)} & T3 ($\uparrow$) \\ \midrule
\multirow{2}{*}{WavLM-large} & Female & 0.961 $\pm$ 0.022 & 0.825 $\pm$ 0.083 & 0.396 $\pm$ 0.073 \\
 & Male & 0.949 $\pm$ 0.029 & 0.662 $\pm$ 0.032 & 0.337 $\pm$ 0.021 \\ \midrule
\multirow{2}{*}{wav2vec2-base} & Female & 0.764 $\pm$ 0.084 & 0.539 $\pm$ 0.079 & 0.305 $\pm$ 0.044 \\
 & Male & 0.798 $\pm$ 0.082 & 0.637 $\pm$ 0.024 & 0.335 $\pm$ 0.041 \\ \midrule
\multirow{2}{*}{wav2vec2-large} & Female & 0.902 $\pm$ 0.058 & 0.734 $\pm$ 0.131 & 0.327 $\pm$ 0.066 \\
 & Male & 0.872 $\pm$ 0.039 & 0.618 $\pm$ 0.042 & 0.348 $\pm$ 0.055 \\ \midrule
\multirow{2}{*}{Hubert-large} & Female & 0.948 $\pm$ 0.017 & 0.634 $\pm$ 0.000 & 0.346 $\pm$ 0.047 \\
 & Male & 0.923 $\pm$ 0.037 & 0.571 $\pm$ 0.000 & 0.322 $\pm$ 0.026 \\ \midrule
\multirow{2}{*}{Whisper-GPT-4o$\ast$} & Female & 0.326 $\pm$ 0.016 & \multicolumn{1}{c}{0.267 $\pm$ 0.015} & 0.242 $\pm$ 0.015 \\
 & Male & 0.396 $\pm$ 0.017 & \multicolumn{1}{c}{0.396 $\pm$ 0.018} & 0.285 $\pm$ 0.016 \\ \bottomrule
\end{tabular}%
}
\label{tab:gender}
\end{table}

\subsection{Evaluation Metrics}
To evaluate the performance of our classification models, we use the Macro F1 score to account for class imbalance, and the Macro Recall score to account for the desired minimization of false negatives in the medical domain. To gauge \ac{ASR} performance, we report the Exact Match (EM), F1-Score, Word Error Rate (WER), Character Error Rate (CER), Word Information Preserved (WIP) and Match Error Rate (MER). EM and F1 show how exactly or how much overlap to the \ac{SLP} professional transcription there is, whereas the WER, WIP, CER and MER provide a standard way to score \ac{ASR} systems.

\begin{table*}[!ht]
\caption{Results of the ablation study (upper), where no hierarchical classification is performed, in comparison to the SOTA combined accross models (lower). Columns show the Macro F1 and Macro Recall score. Results for T1 are equivalent to those in \autoref{tab:main}.}
\centering
\begin{tabular}{lcccc}
\toprule
Model & \multicolumn{2}{c}{T2} & \multicolumn{2}{c}{T3} \\
 & F1 ($\uparrow$) & Recall ($\uparrow$) & F1 ($\uparrow$) & Recall ($\uparrow$) \\ \midrule
WavLM-large & {\ul 0.557 $\pm$ 0.049} & {\ul 0.585 $\pm$ 0.028} & \textbf{0.320 $\pm$ 0.043} & \textbf{0.347 $\pm$ 0.024} \\
wav2vec2-base & 0.483 $\pm$ 0.144 & 0.544 $\pm$ 0.101 & {\ul 0.293 $\pm$ 0.034} & 0.310 $\pm$ 0.038 \\
wav2vec2-large & 0.349 $\pm$ 0.138 & 0.425 $\pm$ 0.124 & 0.255 $\pm$ 0.039 & 0.293 $\pm$ 0.021 \\
Hubert large & \textbf{0.599 $\pm$ 0.075} & \textbf{0.596 $\pm$ 0.065} & 0.286 $\pm$ 0.030 & {\ul 0.323 $\pm$ 0.017} \\ \midrule
Best from \cite{patel2025sound} & 0.318 $\pm$ 0.017 & - & 0.222 $\pm$ 0.015 & - \\ \bottomrule
\end{tabular}%
\label{tab:ablation}
\end{table*}

\subsection{Ablation study}
To understand if the hierarchical classification is beneficial, we perform an ablation study in which we approach each task as a separate classification problem, training a separate model for each task. In this setting, we do not use the cascading mechanism. To enable comparability, we use the same hyperparameter tuning and augmentations as for the cascading approach.

\subsection{Hyperameter Search}\label{sec-hyperparams}

For the \ac{ASR} model, the hyperparameter search space includes learning rates $\eta \in [1 \times 10^{-5}, 5 \times 10^{-4}]$, LoRA ranks $r \in \{64, 96, 128\}$, and LoRA dropout rates $\in \{0.0, 0.1, 0.15, 0.2, 0.3\}$. The \ac{ASR} data augmentation parameters sweep over noise probabilities $p_N \in \{0.4, 0.5, 0.6, 0.7, 0.8\}$, maximum noise amplitudes $\in \{0.025, 0.035, 0.04, 0.05\}$, and maximum pitch shifts $\in \{4, 6, 8, 10\}$ semitones. For tasks T1-3, we optimize learning rates $\eta \in [10^{-5}, 10^{-3}]$, gradient accumulation steps $\in \{2, 4, 6\}$, and oversampling multipliers $M_{os} \in \{1, 3, 5, 8\}$. The T1-3 audio augmentation search covers pitch shift probabilities $p_G \in \{0.2, 0.3, 0.5\}$, with minimum and maximum pitch shifts chosen from $\{0, 1, 3, 4\}$ and $\{0, 4, 6, 8\}$ semitones, respectively.

\subsection{Automatic Speech Recognition}
For the \ac{ASR}-task we investigate three models: whisper-large-v3-turbo, whisper-large-v3 and whisper-large-v2. These were fine-tuned using LoRa-finetuning to transcribe the pathological speech as closely to the speech pathologist ground-truth labels as possible. The training-set was augmented with the augmentations described in §\ref{sec-augmentation}. Inspecting the dataset, not all ground truths are pure transcriptions, but also include vowel shape descriptions, such as ''close-front-round-vowel post-test no-context''. We run the evaluation both with these samples and without to allow for comparability to previous works, but to also provide scores on pure transcriptions. We use the full-length speech as input for \ac{ASR}.

\section{Results and Discussion}

\subsection{Classification Tasks}
Addressing \textbf{RQ1}, our results show that the hierarchical classification pipeline with \ac{SRM} consistently outperforms the current \ac{SOTA} that is based on (multimodal) \ac{LLM} as seen in \autoref{tab:main}.

On $T1$, our best model, WavLM-large, improves over the SOTA \acp{LLM} with a F1-Score of $0.956$ compared to 0.535. It can be seen that all of the classifiers, are better suited for the task with even the lowest scoring model improving by $0.262$ over the current SOTA. 
Furthermore, in response to \textbf{RQ2}, we find that leveraging the hierarchical annotation structure successfully improves classification across the more granular tasks. By utilizing a cascading pipeline, the models are able to predict the specific type of \ac{SLP} ($T2$) to a much higher degree, achieving an F1-score of 0.697 (an increase of 0.379 over the \ac{SOTA}). While performance for the most-granular level of symptom classification ($T3$) is lower, our method still achieves a relevant improvement  (0.354 vs. 0.222). It is important to note that the lower absolute performance on $T3$ may be influenced by a severe class imbalance, which we partially mitigate with oversampling and loss weighting. For instance, the original dataset contains 523 samples of typical speech compared to a mere 29 samples for certain symptoms, representing just 3\,\% of the entire dataset, making training significantly more difficult. 

\subsection{Automatic Speech Recognition}
For a fair comparison with the SOTA established by Patel et al.~\cite{patel2025sound}, we evaluate models on the dataset with clinical vowel descriptions first. Here, our Whisper models outperformed their fine-tuned Qwen2-Audio-7B-Instruct (cf. \autoref{tab:ASR}). Aligning with Ahn et al.~\cite{ahn2025automatic}, our intuition is that \acp{LLM} tend to over-correct pathological speech to form grammatically correct sentences and orthographically correct words, rather than accurately transcribing the exact vocalizations.
Removing vowel shape descriptions to evaluate pure speech transcription improved model performance even further. Here, Whisper-large-v3-turbo achieved 0.640 EM, 0.814 F1, and 0.194 WER. Notably, despite having roughly half the parameters of v2, v3-turbo performs comparably, making it highly advantageous for resource-constrained clinical settings.

\subsection{Effect of Data Augmentation}

Addressing \textbf{RQ3}, we show that using a gender-based pitch augmentation effectively mitigates the gender bias, as seen in \autoref{tab:gender}.  The performance of the models is much fairer between the genders across all three classification tasks, compared to~\cite{patel2025sound}.  

\subsection{Ablation study}
Our ablation study shows that each of the downstream tasks can be improved greatly by employing the hierarchical pipeline. Leveraging the strong binary classifier allows the downstream classifiers to focus exclusively on learning the subtle differences between specific pathologies. Every single \ac{SRM} benefited from this approach. The top-performing model, WavLM-Large, saw its Macro F1-score improve by an absolute margin of 0.140 for $T2$ and 0.034 for $T3$. We observed consistent F1-score improvements across the other architectures as well, including wav2vec2-base (+0.107 for $T2$; +0.043 for $T3$), wav2vec2-large (+0.297 for $T2$; +0.096 for $T3$), and Hubert-Large (+0.024 for $T2$; +0.047 for $T3$). Interestingly, our ablation results reveal that even without the cascading mechanism, the \acp{SRM}  still outperform the current SOTA (multimodal) \ac{LLM} approaches. Thus, the ablation study confirms \textbf{RQ2}, and reinforces the positive answer to \textbf{RQ1}.  

\section{Conclusion}
Our study found that \ac{SRM} still outperform \ac{LLM} for \ac{SSD} detection. We improve upon the state of the art in both \ac{SSD} detection and \ac{ASR} of disordered speech on the SLPHelmUltraSuitePlus~\cite{patel2025sound}  benchmark and publish our code and trained models.
Moreover, we find that a hierarchical classification pipeline further improves the performance of \ac{SLP} tasks on fine-grained annotation. 
Thus, our results act as an important reminder that \acp{LLM} should not be used naively, without also evaluating typical approaches.
Nevertheless, our study has some limitations. First, we focus on a single benchmark, which, to the best of our knowledge, is the only one providing such fine-grained annotations. Next, we trim the audio to twelve seconds for computational reasons. Using the full length might lead to further improvements.
Future research should continue to explore robust data augmentation to mitigate biases and improve performance for clinical \ac{SLP} tasks.

{
\section{Generative AI Use Disclosure} 
Generative AI was used for checking grammar as well as sentence structuring. We are solely responsible and accountable for the quality and content of this work.}

\bibliographystyle{IEEEtran}
\bibliography{mybib}

@inproceedings{patel2025sound,
  title={The Sound of Syntax: Finetuning and Comprehensive Evaluation of Language Models for Speech Pathology},
  author={Patel, Fagun and Nguyen, Duc Quang and Truong, Sang T and Vaynshtok, Jody and Koyejo, Sanmi and Haber, Nick},
  booktitle={Proceedings of the 2025 Conference on Empirical Methods in Natural Language Processing},
  pages={34895--34913},
  year={2025}
}

@article{wav2vec,
  title={wav2vec 2.0: A framework for self-supervised learning of speech representations},
  author={Baevski, Alexei and Zhou, Yuhao and Mohamed, Abdelrahman and Auli, Michael},
  journal={Advances in neural information processing systems},
  volume={33},
  pages={12449--12460},
  year={2020}
}

@article{hubert,
  title={Hubert: Self-supervised speech representation learning by masked prediction of hidden units},
  author={Hsu, Wei-Ning and Bolte, Benjamin and Tsai, Yao-Hung Hubert and Lakhotia, Kushal and Salakhutdinov, Ruslan and Mohamed, Abdelrahman},
  journal={IEEE/ACM transactions on audio, speech, and language processing},
  volume={29},
  pages={3451--3460},
  year={2021},
  publisher={IEEE}
}

@article{wavlm,
  title={Wavlm: Large-scale self-supervised pre-training for full stack speech processing},
  author={Chen, Sanyuan and Wang, Chengyi and Chen, Zhengyang and Wu, Yu and Liu, Shujie and Chen, Zhuo and Li, Jinyu and Kanda, Naoyuki and Yoshioka, Takuya and Xiao, Xiong and others},
  journal={IEEE Journal of Selected Topics in Signal Processing},
  volume={16},
  number={6},
  pages={1505--1518},
  year={2022},
  publisher={IEEE}
}

@article{phi4,
  title={Phi-4-mini technical report: Compact yet powerful multimodal language models via mixture-of-loras},
  author={Abouelenin, Abdelrahman and Ashfaq, Atabak and Atkinson, Adam and Awadalla, Hany and Bach, Nguyen and Bao, Jianmin and Benhaim, Alon and Cai, Martin and Chaudhary, Vishrav and Chen, Congcong and others},
  journal={arXiv preprint arXiv:2503.01743},
  year={2025}
}

@misc{gpt4o,
  title = {{{GPT-4o System Card}}},
  author = {OpenAI},
  year = 2024,
  month = aug,
  howpublished = {https://openai.com/index/gpt-4o-system-card/},
}

@inproceedings{whisper,
  title={Robust speech recognition via large-scale weak supervision},
  author={Radford, Alec and Kim, Jong Wook and Xu, Tao and Brockman, Greg and McLeavey, Christine and Sutskever, Ilya},
  booktitle={International conference on machine learning},
  pages={28492--28518},
  year={2023},
  organization={PMLR}
}

@article{shriberg1999prevalence,
  title={Prevalence of speech delay in 6-year-old children and comorbidity with language impairment},
  author={Shriberg, Lawrence D and Tomblin, J Bruce and McSweeny, Jane L},
  journal={Journal of speech, language, and hearing research},
  volume={42},
  number={6},
  pages={1461--1481},
  year={1999},
  publisher={American Speech-Language-Hearing Association}
}

@article{black2015communication,
  title={Communication Disorders and Use of Intervention Services among Children Aged 3-17 Years: United States, 2012. NCHS Data Brief. Number 205.},
  author={Black, Lindsey I and Vahratian, Anjel and Hoffman, Howard J},
  journal={Centers for Disease Control and Prevention},
  year={2015},
  publisher={ERIC}
}

@inproceedings{hitchcock2015social,
  title={Social, emotional, and academic impact of residual speech errors in school-aged children: A survey study},
  author={Hitchcock, Elaine R and Harel, Daphna and Byun, Tara McAllister},
  booktitle={Seminars in speech and language},
  volume={36},
  number={04},
  pages={283--294},
  year={2015},
  organization={Thieme Medical Publishers}
}

@article{mccormack2009systematic,
  title={A systematic review of the association between childhood speech impairment and participation across the lifespan},
  author={McCormack, Jane and McLeod, Sharynne and McAllister, Lindy and Harrison, Linda J},
  journal={International Journal of Speech-Language Pathology},
  volume={11},
  number={2},
  pages={155--170},
  year={2009},
  publisher={Taylor \& Francis}
}

@article{foster2023speech,
  title={Speech-language disorder severity, academic success, and socioemotional functioning among multilingual and English children in the United States: The National Survey of Children’s Health},
  author={Foster, Matthew E and Choo, Ai Leen and Smith, Sara A},
  journal={Frontiers in Psychology},
  volume={14},
  pages={1096145},
  year={2023},
  publisher={Frontiers Media SA}
}

@article{daniel2017children,
  title={Children with speech sound disorders at school: Challenges for children, parents and teachers},
  author={Daniel, Graham R and McLeod, Sharynne},
  journal={Australian Journal of Teacher Education (Online)},
  volume={42},
  number={2},
  pages={81--101},
  year={2017}
}

@article{jones2005randomised,
  title={Randomised controlled trial of the Lidcombe programme of early stuttering intervention},
  author={Jones, Mark and Onslow, Mark and Packman, Ann and Williams, Shelley and Ormond, Tika and Schwarz, Ilsa and Gebski, Val},
  journal={bmj},
  volume={331},
  number={7518},
  pages={659},
  year={2005},
  publisher={British Medical Journal Publishing Group}
}

@article{mcis1998effectiveness,
  title={Effectiveness of speech intervention for phonological disorders: A randomized controlled trial},
  author={Deborah Almost and Peter Rosenbaum},
  journal={Developmental Medicine \& Child Neurology},
  volume={40},
  number={5},
  pages={319--325},
  year={1998},
  publisher={Wiley Online Library}
}

@article{asha2024,
  title = {2024 {{Schools Survey}}: {{SLP Caseload}} and {{Workload Characteristics}}},
  author = {{American Speech-Language-Hearing Association}},
  year = 2024,
  langid = {english}
}

@article{ashaSchool2024,
  title = {2024 {{Schools Survey}}: {{SLP Workforce}} and {{Work Conditions}}},
  author = {{American Speech-Language-Hearing Association}},
  year = 2024,
  langid = {english}
}

@misc{hu2021loralowrankadaptationlarge,
      title={LoRA: Low-Rank Adaptation of Large Language Models}, 
      author={Edward J. Hu and Yelong Shen and Phillip Wallis and Zeyuan Allen-Zhu and Yuanzhi Li and Shean Wang and Lu Wang and Weizhu Chen},
      year={2021},
      eprint={2106.09685},
      archivePrefix={arXiv},
      primaryClass={cs.CL},
      url={https://arxiv.org/abs/2106.09685}, 
}

@inproceedings{kim2024automatic,
  title={Automatic Children Speech Sound Disorder Detection with Age and Speaker Bias Mitigation.},
  author={Kim, Gahye and Eom, Yunjung and Sung, Selina S and Ha, Seunghee and Yoon, Tae-Jin and So, Jungmin},
  booktitle={Interspeech},
  year={2024}
}

@inproceedings{getman2022wav2vec2,
  title={Wav2vec2-based speech rating system for children with speech sound disorder},
  author={Getman, Yaroslav and Al-Ghezi, Ragheb and Voskoboinik, Katja and Gr{\'o}sz, Tam{\'a}s and Kurimo, Mikko and Salvi, Giampiero and Svendsen, Torbj{\o}rn and Str{\"o}mbergsson, Sofia},
  booktitle={Interspeech},
  year={2022},
  organization={International Speech Communication Association}
}

@article{ahn2025automatic,
  title={Automatic speech recognition (ASR) for the diagnosis of pronunciation of speech sound disorders in Korean children},
  author={Ahn, Taekyung and Hong, Yeonjung and Im, Younggon and Kim, Do Hyung and Kang, Dayoung and Jeong, Joo Won and Kim, Jae Won and Kim, Min Jung and Cho, Ah-Ra and Nam, Hosung and others},
  journal={Clinical linguistics \& phonetics},
  volume={39},
  number={10},
  pages={913--926},
  year={2025},
  publisher={Taylor \& Francis}
}

@incollection{marx2025automatic,
  title={Automatic detection of speech sound disorders in German-speaking children: augmenting the data with typically developed speech},
  author={Marx, Darline Monika and Matassoni, Marco and Brutti, Alessio and others},
  booktitle={Proceedings of Interspeech 2025},
  pages={2875--2879},
  year={2025}
}

@inproceedings{macdonald2021disordered,
  title={Disordered Speech Data Collection: Lessons Learned at 1 Million Utterances from Project Euphonia.},
  author={MacDonald, Robert L and Jiang, Pan-Pan and Cattiau, Julie and Heywood, Rus and Cave, Richard and Seaver, Katie and Ladewig, Marilyn A and Tobin, Jimmy and Brenner, Michael P and Nelson, Philip C and others},
  booktitle={Interspeech},
  volume={2021},
  pages={4833--4837},
  year={2021}
}

@inproceedings{eshkyUltraSuiteRepositoryUltrasound2018,
  title = {{{UltraSuite}}: {{A Repository}} of {{Ultrasound}} and {{Acoustic Data}} from {{Child Speech Therapy Sessions}}},
  shorttitle = {{{UltraSuite}}},
  booktitle = {Interspeech 2018},
  author = {Eshky, Aciel and Ribeiro, Manuel Sam and Cleland, Joanne and Richmond, Korin and Roxburgh, Zoe and Scobbie, James M and Wrench, Alan},
  year = 2018,
  month = sep,
  pages = {1888--1892},
  publisher = {ISCA},
  doi = {10.21437/Interspeech.2018-1736},
  langid = {english}
}

@inproceedings{ngAutomaticDetectionSpeech2022,
  title = {Automatic {{Detection}} of {{Speech Sound Disorder}} in {{Child Speech Using Posterior-based Speaker Representations}}},
  booktitle = {Interspeech 2022},
  author = {Ng, Si-Ioi and Ng, Cymie Wing-Yee and Wang, Jiarui and Lee, Tan},
  year = 2022,
  month = sep,
  pages = {2853--2857},
  publisher = {ISCA},
  doi = {10.21437/Interspeech.2022-935},
  langid = {english}
}

@inproceedings{hansenBestPracticesConsiderations2025,
  title = {Best {{Practices}} and {{Considerations}} for {{Child Speech Corpus Collection}} and {{Curation}} in {{Educational}}, {{Clinical}}, and {{Forensic Scenarios}}},
  booktitle = {10th {{Workshop}} on {{Speech}} and {{Language Technology}} in {{Education}} ({{SLaTE}})},
  author = {Hansen, John H.L. and Dutta, Satwik and Grand, Ellen},
  year = 2025,
  month = aug,
  pages = {123--127},
  publisher = {ISCA},
  doi = {10.21437/SLaTE.2025-25},
  langid = {english}
}

@inproceedings{roseroAdvancingPediatricASR2025,
  title = {Advancing Pediatric {{ASR}}: {{The}} Role of Voice Generation in Disordered Speech},
  shorttitle = {Advancing Pediatric {{ASR}}},
  booktitle = {Proc. {{Interspeech}} 2025},
  author = {Rosero, Karen and Salman, Ali N. and Chandra, Shreeram and Sisman, Berrak and Slot, Cortney Van't and Kane, Alex and Hallac, Rami R. and Busso, Carlos},
  year = 2025,
  pages = {2890--2894}
}

@inproceedings{smithImprovingChildSpeech2017,
  title = {Improving {{Child Speech Disorder Assessment}} by {{Incorporating Out-of-Domain Adult Speech}}.},
  booktitle = {Interspeech},
  author = {Smith, Daniel V. and Sneddon, Alex and Ward, Lauren and Duenser, Andreas and Freyne, Jill and {Silvera-Tawil}, David and Morgan, Angela},
  year = 2017,
  pages = {2690--2694}
}

\end{document}